\documentclass[letterpaper]{article}
\usepackage[preprint]{aaai2027}
\usepackage[hyphens]{url}
\usepackage{graphicx}
\usepackage{natbib}
\usepackage{caption}
\usepackage{booktabs}
\usepackage{amsmath}
\usepackage{amssymb}
\title{SIGMA: Semantic Identifier Grouping for Molecular Autoregression}
\author{
Xinyu Wang\textsuperscript{\rm 1},
Fei Dou\textsuperscript{\rm 2},
Jinbo Bi\textsuperscript{\rm 1},
Minghu Song\textsuperscript{\rm 3}
}
\affiliations{
\textsuperscript{\rm 1}School of Computing, University of Connecticut, Storrs, CT, USA\\
\textsuperscript{\rm 2}Institute for Artificial Intelligence, University of Georgia, Athens, GA, USA\\
\textsuperscript{\rm 3}Institute of Hydrobiology, Chinese Academy of Sciences, Wuhan, China
}

\newcommand{\method}{SIGMA}

\begin{document}
\maketitle

\begin{abstract}
Autoregressive molecular models assign probability to molecular serializations
even though chemical identity is invariant to serialization.  Equivalent
serializations can therefore represent a common molecular identity yet induce
inconsistent next-token decisions.  Randomized strings broaden exposure, but
do not reveal which intermediate decisions should agree.  We introduce
	\method{}, a dense suffix-position objective built from chemically certified
same-suffix triplets: two equivalent histories, one non-equivalent history,
and a shared suffix.  SIGMA aligns
corresponding pre-token hidden states along the continuation and separates the negative
to a finite relative margin, leaving the language-model
objective, decoder, and inference procedure unchanged.  We compare SIGMA
	with canonical training, randomized-serialization training, and last-token alignment across
	four datasets under SMILES and SELFIES.  Across the eight
	representation--dataset blocks, SIGMA yields clear test-reference Fr\'echet
	ChemNet Distance reductions in six: all four SELFIES domains and QM9 and ZINC
	under SMILES, with paired 95\% confidence intervals below zero against every
	control.  Position-wise analyses show improved state
	correspondence, chemical discrimination, and next-token agreement while
	preserving between-molecule information.  On two full-corpus ZINC blocks,
    compute-matched ablations identify chemically correct state
    correspondence as the effective ingredient.  Beyond generation, SIGMA
    improves mean predictive performance on all six molecular property
    benchmarks and reduces sensitivity to equivalent molecular serializations
    on every task.
\end{abstract}

\section{Introduction}

The combinatorial scale of drug-like chemical space places exhaustive
experimental enumeration beyond reach, motivating generative models that can
propose molecular candidates beyond observed libraries
\citep{reymond2012enumeration,ruddigkeit2012enumeration,
polykovskiy2020molecular}.  Autoregressive models provide a scalable approach
by factorizing molecular generation into next-token decisions over serialized
graphs.  This convenience, however, introduces a structural mismatch: the
model assigns probability in string space, whereas chemical identity is
defined by the underlying molecular graph and is invariant to serialization.
SMILES admits different graph traversals \citep{weininger1988smiles}, SELFIES
changes the grammar but retains alternative valid orders \citep{krenn2020self},
and SAFE decomposes molecules into fragments and attachment identifiers
\citep{noutahi2024safe}.  The resulting many-to-one map from strings to
molecules permits distinct serialization histories to denote a common molecular identity.
For example, \texttt{CCO} and \texttt{C(C)O} are alternative SMILES for
ethanol.  Maximum-likelihood training remains correct for each string
individually, but imposes no relation between chemically equivalent
serialization histories---even when they have reached the same molecular
state and face an identical continuation.

\begin{figure*}[t]
  \centering
  \includegraphics[width=.94\textwidth]{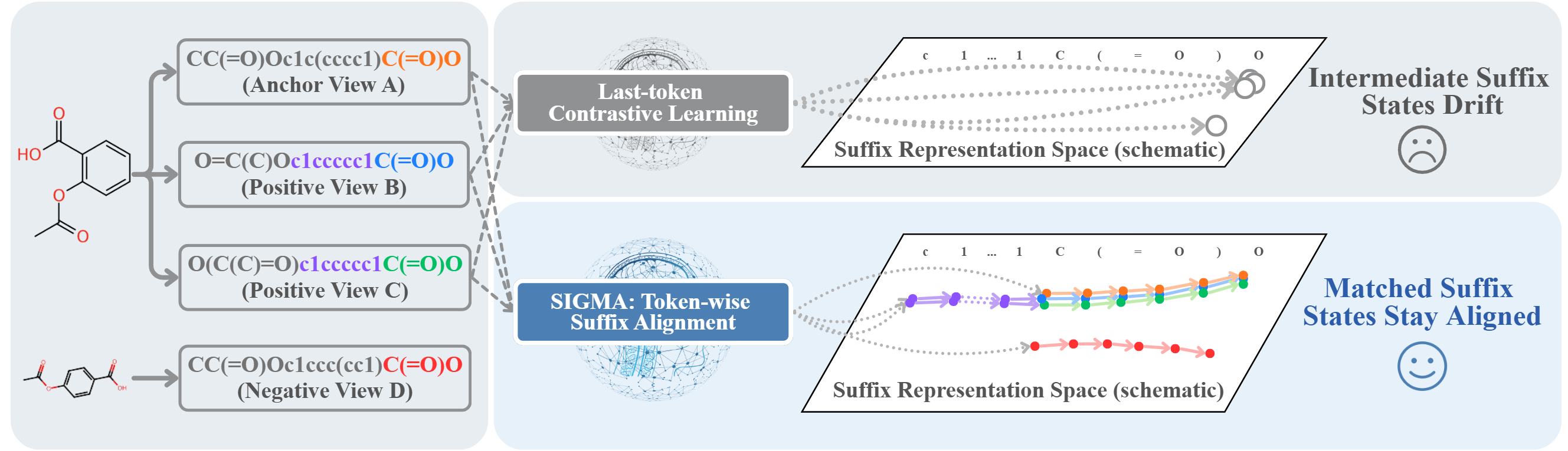}
  \caption{Last-token alignment constrains only the endpoint, whereas SIGMA
  aligns every position of an exact shared suffix and retains a margin-gated
  non-equivalent history.  The artwork uses positive B for the upper example
  and positive C for the lower example; these are alternative illustrations,
  not simultaneous inputs.  Each implemented triplet contains one anchor, one
  equivalent positive, and one non-equivalent negative.  Artwork D therefore
  corresponds to the negative denoted C in the Method notation.}
  \label{fig:sigma-method}
\end{figure*}

The mismatch matters at the level where an autoregressive model acts: the
probability distribution for its next token.  Chemically equivalent molecular
descriptions can elicit inconsistent internal representations or downstream
predictions \citep{ganeeva2024lost,yan2025inconsistency}.  Randomized molecular strings
are the standard remedy \citep{bjerrum2017smiles,arus2019randomized}; they
improve coverage and form a strong baseline, but coverage is not
correspondence.  Exposure to two complete strings for one molecule does not
identify which intermediate states correspond.  Aligning only their final
states has the same limitation: it leaves the ordered pre-token hidden states along
the continuation---the \emph{suffix prediction path}---unconstrained.  The
desired invariance is therefore conditional:
when two histories encode the same chemistry and face the same continuation,
their decisions should agree at each continuation token, while a chemically
different history should remain distinguishable.

We introduce \method{}---\textbf{S}emantic \textbf{I}dentifier
\textbf{G}rouping for \textbf{M}olecular \textbf{A}utoregression---which turns
serialization ambiguity into certified supervision.  SIGMA constructs two
equivalent histories $A$ and $B$ and one length-controlled, valid but
non-equivalent history $C$, all followed by exactly the same suffix.  We call
$B$ the positive and $C$ the negative only as concise loss notation; neither
term denotes a property label or an invalid molecule.  Each position in
that suffix is then a controlled event: all three serialization histories are about to
predict the same literal token, but only the positive pair represents the same
chemistry.  At every such position, a position-matching objective requires the anchor
to identify the corresponding state in the equivalent history.  A relative
margin simultaneously requires the positive similarity to exceed the negative
similarity, with no further repulsion after the margin is satisfied.

Figure~\ref{fig:sigma-method} contrasts this dense suffix-position supervision with
endpoint-only contrast: the latter leaves intermediate suffix states
unconstrained, whereas SIGMA aligns every certified suffix decision and
retains a margin-gated non-equivalent history.

SIGMA augments ordinary next-token training without changing the
language-model head, decoder, or inference cost.  Prefixes are aligned by
position within the shared suffix rather than absolute sequence index, because
the two serialization histories can have different lengths.  Padding is removed before the
position-identification loss is normalized, and the negative contributes no
gradient after the desired ordering is achieved.  Independent checks verify
the positive identity, negative non-identity, literal suffix equality,
next-token timing, and padding exclusion.  This construction makes the
mechanism testable: the model should identify corresponding positive states
more accurately, rank negatives below them, and produce more similar
next-token distributions without erasing distinctions between different
molecules.

We make four contributions:
\begin{itemize}
    \item We formulate \emph{serialization-induced prediction inconsistency}:
    equivalent molecular strings can induce incompatible intermediate
    next-token decisions.  This separates explicit state correspondence from
    string enumeration.
    \item We introduce chemically certified same-suffix triplets and a
    dense suffix-position objective with a margin-gated non-equivalent history.
    \item We compare canonical training, randomized-serialization training,
    last-token alignment, and SIGMA under matched data, initialization,
    architecture, optimization, and decoding within each SMILES/SELFIES
    representation--dataset block; SAFE is used to
    analyze a fragment-assembly limitation.
    \item We evaluate both generated distributions and the proposed mechanism:
    correct-position accuracy, ordering-violation rate, next-token agreement,
    suffix-token negative log-likelihood (NLL), and information-preservation
    checks.
    Full-corpus controlled comparisons distinguish correct hidden correspondence
    from a prespecified symmetric-KL output-space variant, shuffled-pair
    computation, and extra compute.
    A secondary whole-molecule study tests the same equivalence principle on
    six property benchmarks.
\end{itemize}
\section{Related Work}

\paragraph{Chemical language models and molecular strings.}
Transformer backbones
\citep{vaswani2017attention,devlin2019bert,radford2019language,
liu2019roberta,jiang2023mistral} underpin chemical models for representation
learning and property prediction
\citep{chithrananda2020chemberta,wang2019smiles,honda2019smiles,
fabian2020molecular}, autoregressive design
\citep{bagal2021molgpt,li2023druggpt}, and tool use
\citep{bran2024chemcrow}.  Training resources range from PubChem, ChEMBL, and
ZINC corpora \citep{wang2009pubchem,gaulton2012chembl,irwin2005zinc,
sterling2015zinc} to graph and multimodal biological views
\citep{xu2024triple,seal2022integrating,seal2023merging}.  SMILES admits multiple
valid traversals \citep{weininger1988smiles}; DeepSMILES revises ring/branch
syntax \citep{o2018deepsmiles}, SELFIES constrains the grammar
\citep{krenn2020self}, and t-SMILES/SAFE use fragments
\citep{wu2024t,noutahi2024safe}.  These schemes do not themselves identify
which states in two serializations face the same next chemical action.

\paragraph{Order augmentation and molecular language models.}
Randomized SMILES can improve molecular generative models by exposing more
traversals \citep{bjerrum2017smiles,arus2019randomized}; related augmentation
is also effective in reaction modeling \citep{tetko2020state}.  All-SMILES
aggregates several complete serializations \citep{alperstein2019all}.
Equivalent descriptions can nevertheless elicit inconsistent representations
or task predictions \citep{ganeeva2024lost,yan2025inconsistency}, while strong
point predictions need not imply reliable uncertainty estimates
\citep{scalia2020evaluating}.  Enumeration teaches which strings occur, not
which internal states face the same continuation; randomized serialization is
therefore a strong exposure baseline, not semantic alignment.

\paragraph{Molecular generation.}
Graph generators construct molecules through graph-space actions
\citep{you2018graph,shi2020graphaf}, supported by message-passing
representations \citep{gilmer2017neural,hu2020strategies};
reinforcement learning optimizes task rewards under sample-efficiency
constraints \citep{olivecrona2017molecular,gao2022sample}, and fragment
retrieval augments generation \citep{lee2024molecule}.  Learning-order models learn
a state-dependent graph-generation order \citep{wang2025learning}.  SIGMA
addresses a complementary degree of freedom: it keeps the string model,
generation order, and decoder fixed, then supervises equivalent histories
when they face a common continuation.

\paragraph{Contrastive sequence learning.}
Contrastive learning commonly makes two augmented views agree at a pooled
representation \citep{chen2020simple}; molecular approaches similarly compare
complete strings, graphs, or other global views
\citep{wang2022molecular,pinheiro2022smiclr,qian2024consmi,zhu2023dual}.
Token-aware contrastive objectives have also been used to shape individual
language states rather than only pooled embeddings
\citep{su2022tacl,zhou2024token}.  At the same token-step granularity,
PSV-PPO uses a partial-SMILES validity table to reshape entropy and
distribution-matching losses at every autoregressive step, providing feedback
before completion \citep{wang2025partialsmiles}.  SIGMA instead aligns
chemically equivalent histories and separates non-equivalent histories under
an exact shared continuation.  The shared continuation supplies a controlled
local probe before every corresponding token; last-token alignment tests
sparse endpoint supervision, while controlled ablations isolate position
density from loss form.

\paragraph{Molecular generation evaluation.}
Benchmarks such as MOSES and GuacaMol emphasize that validity alone is
insufficient \citep{polykovskiy2020molecular,brown2019guacamol}.  We therefore
report Fr\'echet ChemNet Distance together with validity and likelihood.
Because distribution fidelity does not verify the proposed mechanism, we
separately test position matching, continuation decisions, and preservation
of between-molecule information.
\section{Method}

\subsection{Equivalent Serialization Histories}

Let $x=(x_1,\ldots,x_T)$ be a molecular string and
$p_\theta(x_t\mid x_{<t})$ an autoregressive model.  We use the
token-normalized language-model (LM) loss implemented by the trainer,
\begin{equation}
  \mathcal{L}_{\mathrm{LM}}
  =
  \frac{
    \sum_i\sum_{t=1}^{T_i}
    -\log p_\theta(x_{i,t}\mid x_{i,<t})
  }{
    \sum_i T_i
  } .
  \label{eq:token-normalized-lm}
\end{equation}
This objective is defined on strings.  It therefore gives no relation between
two token histories assigned the same molecular identity.

\paragraph{Illustrative construction.}
Consider
$A=\texttt{CCO}$ and $B=\texttt{C(C)O}$, two SMILES histories that encode the
same molecule, and $C=\texttt{CCC}$, a valid history that encodes a different
molecule.  We
append the same two-token continuation $S=\texttt{CC}$ to all three.  The
equivalent completions \texttt{CCOCC} and \texttt{C(C)OCC} still identify the
same molecule, whereas the non-equivalent completion \texttt{CCCCC} does not.
This tuple supplies aligned supervision along the two equivalent histories
while retaining a valid, chemically distinct history.  We use it throughout the
section; the constructor retains it only after the chemical checks below pass.

SIGMA formalizes this controlled comparison as follows.  Let
$A$, $B$, and $C$ be prefixes and let
$S=(s_1,\ldots,s_m)$ be a nonempty, token-identical continuation, where
$\oplus$ denotes token concatenation.  We retain
$(A,B,C,S)$ only when $A$ and $B$ remain chemically equivalent while consuming
$S$, whereas $C$ remains non-equivalent under that same continuation.  Hence
the histories differ, but the future prediction events are held fixed.  This
distinguishes SIGMA from view augmentation, which exposes alternate strings
without identifying which intermediate decisions should agree.

For a decoded string $x$, $\operatorname{id}_{\mathrm{std}}(x)$ is its complete
Standard InChIKey, computed with RDKit after a canonical isomeric-SMILES round
trip.  This is the operational identity relation used throughout the paper.
It is reproducible but deliberately coarser than tautomer-specific graph
identity: Standard InChI normalizes mobile hydrogens and some
protonation/charge arrangements because it omits the Fixed-H layer
\citep{heller2015inchi,inchiTrustTechnicalFAQ}.  Accordingly, every
equivalence and collision statement below is relative to this declared
Standard-InChIKey relation.  For the fixed continuation $S$,
\[
 A\equiv_S B
 \quad\Longleftrightarrow\quad
 \operatorname{id}_{\mathrm{std}}(A\!\oplus\!S)
 =\operatorname{id}_{\mathrm{std}}(B\!\oplus\!S),
\]
subject to the position-wise chemical verification below.  To measure whether
equivalent histories
actually make similar decisions, let
$\pi^X_j=p_\theta(\cdot\mid X\!\oplus\!s_{<j})$:
\begin{equation}
\mathcal D_\theta(A,B;S)
=\frac{1}{m}\sum_{j=1}^{m}
\operatorname{JS}(\pi^A_j,\pi^B_j).
\label{eq:decision-inconsistency}
\end{equation}
Here $\operatorname{JS}$ is the Jensen--Shannon divergence between the two
next-token distributions; zero means that they make identical probabilistic
predictions at that suffix position.
SIGMA does not optimize Eq.~\eqref{eq:decision-inconsistency}; it constrains
the pre-token hidden states that parameterize these conditionals and treats reduced
decision inconsistency as a falsifiable mechanism prediction.

\subsection{Chemically Certified Same-Suffix Triplets}

The complete fail-closed construction is given in Supplementary
Algorithm~1, Section~3.1 (``Triplet Construction Algorithm'').  It checks
lossless tokenization, matched parseability,
position-wise identities, exact suffix reuse, context length, and the stored
source identity before emitting any auxiliary example.

The search for $C$ uses prefix-token length buckets satisfying
$|\,|C|-|A|\,|\leq\max(3,\operatorname{round}(.25|A|))$.  A hash-seeded
starting point removes first-row bias, after which candidates are checked
exhaustively up to a fixed search limit.  Unlike an incidental in-batch
negative, $C$ is explicitly matched in length and continuation: it is a valid
non-equivalent history evaluated under the exact same suffix.

\paragraph{Illustrative certificate check.}
The complete-molecule identities in the example are:
\begin{quote}
\raggedright
\textbf{Equivalent pair $(A,B)$}\\
\texttt{RTZKZFJDLAIYFH-UHFFFAOYSA-N}\\
\textbf{Non-equivalent history $C$}\\
\texttt{OFBQJSOFQDEBGM-UHFFFAOYSA-N}
\end{quote}
All three histories face the same two literal token events, while only
$(A,B)$ belongs to the same conditional chemical equivalence class.

\subsection{Suffix-Relative Pre-Token Alignment}

The equivalent histories above contain three versus five prefix tokens.
Pairing equal
absolute indices would compare different events and may pair a real state with
padding.  Let $g^X_t$ be the final-layer state at absolute
predictive position $t$, and let $r_X(j)$ map suffix offset $j$ to its
pre-token position in serialization $X$.  SIGMA compares
\begin{align}
  h^A_j &= g^A_{r_A(j)}=H_\theta(A \oplus s_{<j}), \nonumber\\
  h^B_j &= g^B_{r_B(j)}=H_\theta(B \oplus s_{<j}),\\
  h^C_j &= g^C_{r_C(j)}=H_\theta(C \oplus s_{<j}). \nonumber
\end{align}
If the first suffix target in $X$ has index $b_X$, then
$r_X(j)=b_X+j-2$.  In the example, the first suffix token is predicted by
states $(3,5,3)$ for $(A,B,C)$, not a shared absolute index.  Only real suffix
tokens are supervised; \texttt{EOS} is excluded as a target.  Pre-token timing
is essential because a post-token state has already consumed the decision
being compared.

\subsection{Dense Suffix-Position Matching with Margin-Gated Negatives}

The positive term is a position-identification task.  Given the anchor state
in $A$ immediately before suffix token $s_j$, the model must identify which
state in the entire $B$ serialization is immediately before that same token.
Choosing a nearby but causally different position counts as an error.  The
implemented candidate
set $\mathcal V_B$ includes all real states from the start-of-sequence
(\texttt{GO}) through the end-of-sequence (\texttt{EOS}) state and excludes
only right padding; the supervised targets remain the real
suffix-pre-token positions.  With cosine similarity and temperature $\tau$,
\begin{align}
q_\tau(k\mid h^A_j;B)
&=
\frac{\exp(\cos(h^A_j,g^B_k)/\tau)}
{\sum_{\ell\in\mathcal V_B}
 \exp(\cos(h^A_j,g^B_\ell)/\tau)}, \nonumber\\
\ell^+_j
&=-\log q_\tau(r_B(j)\mid h^A_j;B).
\label{eq:directed-position-matching}
\end{align}
Thus $\ell^+_j$ is small only when the correct corresponding position receives
high matching probability.  Because
$|\mathcal V_B|$ changes with sequence length, raw position-matching losses
are not compared across datasets; all method comparisons use identical
triplets within a dataset.  Padding logits are set to $-\infty$ before the
softmax.

The negative term enforces finite relative discrimination: the chemically
different history should be at least $m$ less similar to the anchor
than the equivalent history.  Define the aligned similarities
\[
  s^+_j=\cos(h^A_j,h^B_j),\qquad
  s^-_j=\cos(h^A_j,h^C_j).
\]
The negative term requests only a finite relative ordering,
\begin{equation}
  \ell^{\mathrm{gate}}_j
  =\max(0,s^-_j-s^+_j+m).
  \label{eq:margin-gate}
\end{equation}
It is active while the non-equivalent state is too close and becomes exactly
zero once $s^+_j\geq s^-_j+m$.  Thus SIGMA does not indefinitely repel
states that may still share ordinary chemical features.

The certificate is symmetric in the chemical relation, while the
position-matching loss is directed: the anchor history $A$ queries its equivalent
serialization $B$.

Let $\mathcal I$ contain all retained example--suffix-offset pairs.  We average
$\ell^+_j$ and $\ell^{\mathrm{gate}}_j$ over $\mathcal I$ to obtain
$\mathcal L_+$ and $\mathcal L_{\mathrm{gate}}$.  The total objective is
\begin{equation}
  \mathcal{L}
  =
  \mathcal{L}_{\mathrm{LM}}
  +\lambda_{\mathrm{aux}}\left(
    \mathcal{L}_{+}+\rho_{\mathrm{neg}}\mathcal{L}_{\mathrm{gate}}
  \right).
  \label{eq:full-objective}
\end{equation}
Here $\lambda_{\mathrm{aux}}$ controls the total auxiliary strength and
$\rho_{\mathrm{neg}}$ controls the negative relative to the positive term.
The negative therefore enters the total objective with effective coefficient
$\lambda_{\mathrm{neg}}^{\mathrm{eff}}
=\lambda_{\mathrm{aux}}\rho_{\mathrm{neg}}$.  The shared
validation-selected setting is $\lambda_{\mathrm{aux}}=.075$,
$\rho_{\mathrm{neg}}=1$, $m=.10$, and $\tau=.07$.

\subsection{Controlled Optimization}

We use two fixed-coverage regimes.  In the SELFIES certified-subset regime,
one triplet batch defines an optimizer batch: LM loss is computed on $A$ (or
$B$ for randomized training), while auxiliary methods also forward $B$ and
$C$.  In the SMILES full-corpus regime, every method receives the same
complete-molecule LM schedule and auxiliary methods receive a separately
cycled triplet stream.  Within each representation--dataset block, corpus
coverage, initialization, architecture, optimization, and decoding are
matched.  Micro-batching preserves the token- and position-normalized
objective.  Full execution and training-cost details are supplementary;
checkpoints retain only the unchanged GPT-2 and therefore add no inference
parameters or decoding cost.

\subsection{Why the Objective Has This Form}

The following properties motivate the design; complete statements and proofs
are supplementary.

\paragraph{Well-defined supervision and finite separation.}
Suffix-relative indexing is the unique monotone map from a suffix offset to
the pre-token hidden state that is about to predict that token.  Masking before the
position-matching softmax makes the loss and all real-state gradients exactly invariant
to arbitrary right padding.  The two contrastive terms also have finite,
checkable meanings: high matching probability implies a minimum
target--competitor cosine gap, while
$\ell^{\mathrm{gate}}_j=0$ if and only if
$s^+_j-s^-_j\geq m$.  Thus the negative stops once the requested ordering
is achieved, and complete directional collapse cannot attain zero gated loss
when $m>0$.

\paragraph{Why every continuation position is aligned.}
Let the shared LM head be $z=Wh+b$ and
$M_W=\max_u\|w_u\|_2$.  Evaluating the same continuation $S$ under two
histories gives
\begin{equation}
\left|
\log p_\theta(S\mid A)-\log p_\theta(S\mid B)
\right|
\leq
2M_W\sum_{j=1}^{m}\|h^A_j-h^B_j\|_2 .
\label{eq:continuation-stability}
\end{equation}
Each aligned pre-token hidden state therefore controls one additive term in the
  teacher-forced suffix-token likelihood.  Last-token alignment controls only
one term and leaves the other $m-1$ decisions unrestricted.  Equation
\eqref{eq:continuation-stability} is conditional rather than automatic:
cosine alignment must be accompanied by norm control to bound Euclidean
distance.  We consequently measure next-token JSD, suffix-token NLL,
effective rank, and between-molecule information preservation instead of
treating auxiliary loss as
a behavioral result.

\paragraph{Local versus global scope.}
Certified constraints cover observed histories, not every history reachable
under free generation.  Two models may satisfy every certified constraint yet
differ at an unconstrained history reached with probability $r$, producing
joint variation $r\delta$ when their outgoing conditionals differ by
$\delta$.  This scope result motivates the separate FCD evaluation:
SIGMA is designed to improve local conditional consistency, while the induced
free-generation distribution remains an empirical question.

\subsection{Applying the Verification to SMILES, SELFIES, and SAFE}

The contract is representation-independent, while tokenization and partial
decodability differ.  SMILES uses atomwise tokens and parseable prefixes;
SELFIES uses official bracket tokens and decodable suffix checkpoints; SAFE
retains atomic/basic tokens, dots, and attachment identifiers while checking
fragment and connection boundaries.  Any row that fails its
representation-specific checks is rejected.  Complete rules and counts are
supplementary.
\section{Experimental Setup}

\paragraph{Data, representations, and coverage.}
We evaluate QM9 \citep{ramakrishnan2014quantum}, ZINC
\citep{irwin2005zinc}, ChEMBL \citep{gaulton2012chembl}, and MOSES
\citep{polykovskiy2020molecular}, spanning small organic, purchasable
drug-like, bioactive, and benchmark drug-like molecules.  SMILES and SELFIES
form the primary three-seed study; SAFE provides a focused single-seed
limitation analysis.  Within each representation--dataset block, all methods
share the language-model corpus and initialization.  SMILES uses the complete
parent corpus plus a certified-triplet stream; SELFIES fixes the same
certified-row corpus for all methods.  Contrasts therefore change supervision,
not within-block coverage, although absolute SMILES and SELFIES values are not
a corpus-matched cross-representation comparison.  Exact split, triplet, and
test-reference counts and the independent RDKit/InChI audit
\citep{rdkit2026,heller2015inchi} are supplementary.

\paragraph{Controlled methods and training.}
All methods use the same six-layer, eight-head, 256-dimensional GPT-2,
tokenizer, optimizer schedule, three training epochs, and seeds 101, 202, and
303.  Within each fixed-coverage block they differ only in serialization view
and auxiliary supervision:
\begin{itemize}
  \item \textbf{Canonical} applies ordinary next-token likelihood to the
  designated canonical string $A$.
  \item \textbf{Randomized-serialization training} applies the same likelihood
  to the paired randomized valid serialization $B$, with no alignment loss.
  \item \textbf{Last-token alignment} adds one positive-versus-negative
  comparison at the final real token of the shared suffix.
  \item \textbf{SIGMA} applies positive suffix-position matching and a margin-gated
  negative at every real token of that suffix.
\end{itemize}
The first two test whether alternate-serialization exposure is sufficient; the
third tests a single endpoint constraint.  The validation-selected shared
SIGMA setting uses
$\lambda_{\mathrm{aux}}=.075$, within-auxiliary negative weight
$\rho_{\mathrm{neg}}=1$, and cosine margin $m=.10$.  Hence the effective
positive and gated-negative coefficients are both .075.  SELFIES uses a
method-independent certifiable subset of each formal test split, frozen before
training.  Complete hyperparameters and filtering stages are supplementary.

\paragraph{Evaluation and statistics.}
Fr\'echet ChemNet Distance (FCD; lower is better) is the primary
distribution metric \citep{preuer2018frechet}; validity, diversity,
connectedness, likelihood, and representation probes guard against trivial
improvements or collapse.  Primary results use three training seeds and five
paired generations per seed, reported as mean $\pm$ seed standard deviation
with paired seed-by-generation hierarchical-bootstrap 95\% intervals.  Loss
weights are selected on validation FCD and NLL; reported generations are
scored against frozen formal test references.  The complete metric
definitions, 10,000-draw bootstrap estimand, reference construction, and
per-seed results are given in the supplement.

\paragraph{Molecular property prediction.}
For whole-molecule prediction on six MoleculeNet benchmarks
\citep{wu2018moleculenet}, EOS states of two strings for one molecule are
aligned with a label-aware finite-margin negative.  Validation-selected
checkpoints are evaluated on six property benchmarks against an otherwise
identical randomized-serialization control; task details are supplementary.
\section{Results}

\paragraph{Evidence structure.}
We separate three questions that FCD alone cannot resolve.  First, are the
constructed positives and negatives chemically correct?  Second, does SIGMA
change the intended autoregressive decisions while preserving molecular
distinctions?  Third, do those local changes improve the distribution of
complete molecules?  We report the global outcome first and then establish
the chemical verification, model-side mechanism, and
information-preservation checks that support its interpretation.

\subsection{Generation Fidelity Across Representations}

\begin{table*}[t]
\centering
\begin{small}
\setlength{\tabcolsep}{1.2pt}
\begin{tabular*}{.98\textwidth}{@{\extracolsep{\fill}}lrrrrrrrr@{}}
\toprule
& \multicolumn{4}{c}{SMILES} & \multicolumn{4}{c}{SELFIES}\\
\cmidrule(lr){2-5}\cmidrule(lr){6-9}
Dataset
& Canon. & Random. & Last-token & SIGMA
& Canon. & Random. & Last-token & SIGMA\\
\midrule
QM9
& .6655$\pm$.0610 & .8013$\pm$.0286 & .6474$\pm$.0568 & \textbf{.5832$\pm$.0441}
& .8557$\pm$.0644 & 1.0510$\pm$.1080 & .8491$\pm$.0418 & \textbf{.6846$\pm$.0120}\\
ZINC
& 1.9655$\pm$.2463 & 2.3486$\pm$.2065 & 1.9960$\pm$.2080 & \textbf{1.8348$\pm$.1975}
& 6.8046$\pm$.1125 & 8.1759$\pm$.1164 & 6.9863$\pm$.2009 & \textbf{6.2527$\pm$.1241}\\
ChEMBL
& \textbf{1.5914$\pm$.0851} & 1.7327$\pm$.0958 & 1.6034$\pm$.0999 & 1.6090$\pm$.0879
& 3.1631$\pm$.0683 & 3.7631$\pm$.0756 & 3.2637$\pm$.1064 & \textbf{3.1095$\pm$.0644}\\
MOSES-300k
& 5.7967$\pm$.1501 & 6.1355$\pm$.0387 & 5.8048$\pm$.0690 & \textbf{5.7502$\pm$.0851}
& 9.6606$\pm$.2699 & 11.7648$\pm$.1037 & 9.9398$\pm$.0523 & \textbf{9.1999$\pm$.1524}\\
\bottomrule
\end{tabular*}
\end{small}
\caption{Test-reference FCD under SMILES and SELFIES (mean $\pm$ standard
deviation across three seeds; lower is better; bold is best).}
\label{tab:v41-margin-fcd}
\label{tab:v42-1-selfies-fcd}
\end{table*}

The SMILES half of Table~\ref{tab:v41-margin-fcd} reports the controlled
three-seed comparison against the test generation references.  Relative to the
strongest control mean, SIGMA reduces FCD by 9.9\% on QM9 and
6.7\% on ZINC.  Both reductions hold in all three training seeds, and the
paired interval lies below zero against every control (against canonical:
$\Delta=-.0823$, 95\% CI $[-.1119,-.0554]$ on QM9 and
$\Delta=-.1307$, CI $[-.1909,-.0627]$ on ZINC).  MOSES-300k has the lowest
mean, but its comparisons with canonical training and last-token alignment have
intervals that include zero ($\Delta=-.0466$, CI $[-.1375,+.0553]$ against canonical).
On ChEMBL, SIGMA improves over randomized-serialization training, while its
comparisons with canonical training and last-token alignment include zero.
We therefore claim clear SMILES reductions on QM9 and ZINC; the MOSES-300k and
ChEMBL intervals do not establish a difference from the strongest controls.

Validity improves over canonical training on QM9, ZINC, and MOSES-300k
(.7601$\rightarrow$.8055, .7382$\rightarrow$.7610, and
.8183$\rightarrow$.8315), while decreasing slightly on ChEMBL
(.8723$\rightarrow$.8671).  The FCD gains are therefore not a trivial result
of discarding more invalid outputs.  The complete table also exposes a useful
boundary: local correspondence improves on SMILES--ChEMBL even though its FCD
interval includes zero.  Supplementary Section ``Diagnosing the
SMILES--ChEMBL Boundary'' localizes this mismatch to repeated-token position
competition, ring closure, and stopping calibration.

The SELFIES half of Table~\ref{tab:v42-1-selfies-fcd} repeats the
test-reference comparison.  SIGMA has the lowest mean FCD in every dataset and every training
seed.  Relative to the strongest control, the reductions are 19.4\% on QM9,
8.1\% on ZINC, 1.7\% on ChEMBL, and 4.8\% on MOSES-300k.  Every paired 95\%
interval lies below zero against all three controls.  Against canonical
training, the differences are $-.1712$ $[-.2259,-.1178]$, $-.5519$
$[-.7313,-.3676]$, $-.0536$ $[-.1033,-.0085]$, and $-.4607$
$[-.6451,-.3069]$, respectively.  Thus the clear gains span all four SELFIES
domains rather than one favorable dataset.

\subsection{Consistency Across Suffix Positions}

The validity of the supervision under the declared identity rule is
established on the independently verified QM9 corpus: every retained positive
reconstructs the same Standard-InChIKey equivalence class, every retained
non-equivalent history reconstructs a different class, and all three strings
share the exact suffix.  The independent cross-check found zero
anchor--non-equivalent Standard-InChIKey collisions in that corpus.
This rules out the label-error confound in which the model repels a state that
is equivalent under the certificate.

For every triplet and suffix position we ask four concrete
questions: (i) can the anchor identify the exact corresponding position in the
equivalent serialization; (ii) is the non-equivalent state incorrectly ranked at
least as similar as the equivalent state; (iii) do the equivalent histories
assign similar probabilities to the next token; and (iv) does the model assign
greater likelihood to the literal shared suffix?  Curves across suffix
positions determine whether the effect spans the entire suffix or appears only
near its endpoint.

Concretely, the two primary state decisions at offset $j$ are
\begin{align}
  R^+_j&=\mathbb{1}\!\left[
  \arg\max_{t\in\mathcal V_B}\operatorname{sim}(h^A_j,g^B_t)=r_B(j)\right],
  \nonumber\\
 V^-_j&=\mathbb{1}\!\left[
 \operatorname{sim}(h^A_j,h^C_j)\geq
 \operatorname{sim}(h^A_j,h^B_j)\right].
\end{align}
$R^+_j=1$ means that the anchor identifies the correct position corresponding
to suffix offset $j$; $V^-_j=1$ means that the non-equivalent state is ranked
at least as similar as the equivalent state and therefore violates the desired
ordering.  They are
reported in ten normalized suffix-position bins, so a terminal-only effect
cannot masquerade as dense suffix-position alignment.

\begin{figure}[t]
  \centering
  \includegraphics[width=\columnwidth]{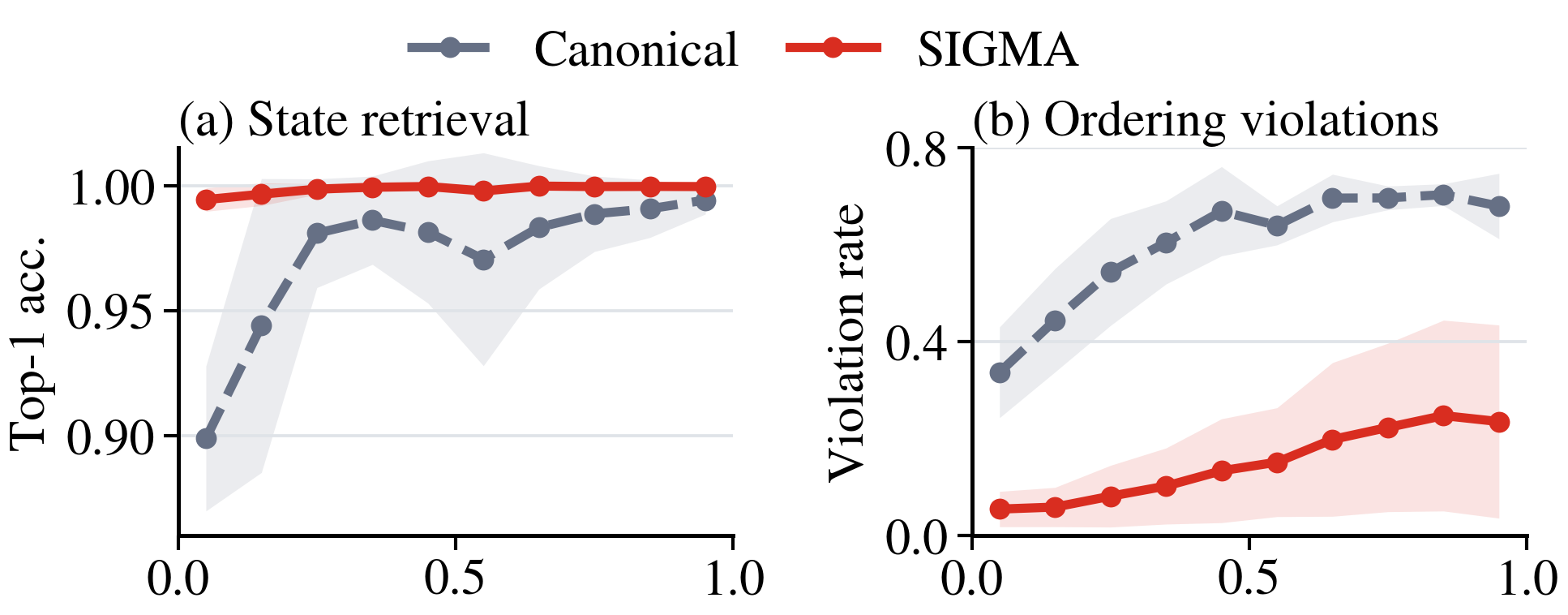}
  \caption{The final margin-gated model changes the whole shared-suffix
  prediction path, not only its endpoint.  Curves aggregate the four SMILES
  datasets and three training seeds (12 dataset--seed conditions); markers are
  condition means and ribbons show one standard deviation across conditions.
  The horizontal axes run from the beginning (0) to the end (1) of the shared
  suffix.  At every normalized suffix bin, SIGMA identifies the aligned state
  of the certified equivalent
  prefix more often and is less likely to rank the certified non-equivalent
  prefix above it.}
  \label{fig:suffix-position-mechanism}
\end{figure}

Figure~\ref{fig:suffix-position-mechanism} shows that both primary effects persist
from the beginning to the end of the shared continuation.  Every listed
SMILES metric moves in the intended direction on every dataset and all three
seeds.  On QM9, for example, the probability that the non-equivalent state outranks the
equivalent state falls from .5405 to .2383, correct-position accuracy
rises from .8875 to .9912, and equivalent-prefix next-token agreement rises
from .7234 to .8014.  The result verifies the proposed local mechanism while
showing, through the blocks without a clear FCD reduction, that improved
suffix-prediction consistency and global distribution fidelity are related but
distinct outcomes.

The SELFIES replication reaches the same mechanism conclusion: on all four
datasets, all six measurements move in the intended direction in all three
seeds, including lower suffix-token NLL.  Supplementary Section~8 reports the
complete cross-domain mechanism table.

\begin{table}[t]
\centering
\begin{small}
\setlength{\tabcolsep}{3.4pt}
\begin{tabular*}{.98\columnwidth}{@{\extracolsep{\fill}}llrr@{}}
\toprule
Task & Metric & \shortstack{Randomized\\serialization} & SIGMA\\
\midrule
ESOL        & RMSE  & $1.013\pm.081$ & $\mathbf{.970\pm.082}$\\
FreeSolv    & RMSE  & $2.320\pm.155$ & $2.246\pm.108$\\
Lipophilicity & RMSE & $.758\pm.021$ & $\mathbf{.722\pm.005}$\\
BACE        & AUROC & $.823\pm.012$ & $.834\pm.013$\\
BBBP        & AUROC & $.920\pm.026$ & $.943\pm.005$\\
ClinTox     & AUROC & $.761\pm.155$ & $.854\pm.061$\\
\bottomrule
\end{tabular*}
\end{small}
\caption{Transfer across six molecular-property benchmarks. Predictions
average two paired serializations per molecule; values are mean $\pm$
training-seed standard deviation. Lower RMSE and higher AUROC are better.
Complete paired 95\% intervals appear in Supplementary Section~5.}
\label{tab:downstream-replay-main}
\end{table}

\paragraph{SAFE boundary.}
In a single-seed SAFE analysis, all local checks improve, but last-token
alignment retains lower overall FCD because dense alignment does not resolve
fragment connectivity.  Conditional on connected outputs, SIGMA improves FCD
on three of four domains.  Supplementary Section~8 reports the complete table
and assembly diagnosis.

\subsection{What Produces the Gain?}

Full-corpus ZINC controls match LM data, optimizer schedule, triplet stream,
and auxiliary compute.  Table~\ref{tab:v48-causal-main} compares a
prespecified symmetric-KL output variant, positive-only
\textbf{SIGMA-Pos}, shuffled-pair computation, and final
\textbf{SIGMA-Margin}.

\begin{table}[t]
\centering
\begin{small}
\setlength{\tabcolsep}{2.4pt}
\begin{tabular*}{.98\columnwidth}{@{\extracolsep{\fill}}lcc@{}}
\toprule
Training objective & S--ZINC & F--ZINC\\
\midrule
Full-corpus LM & 1.094$\pm$.044 & 6.330$\pm$.114\\
Output-distribution alignment & 1.137$\pm$.043 & 6.401$\pm$.226\\
SIGMA-Pos & 1.062$\pm$.069 & \textbf{5.725$\pm$.061}\\
Shuffled-pair compute control & 1.087$\pm$.023 & 6.333$\pm$.068\\
SIGMA-Margin & \textbf{1.043$\pm$.064} & 5.788$\pm$.126\\
\bottomrule
\end{tabular*}
\end{small}
\caption{Separating compute and alignment on two full-corpus ZINC
validation blocks (FCD
$\downarrow$, mean$\pm$SD over three training seeds).  Every row uses the same
full language-model corpus, initialization, optimizer schedule, and five
paired generation repetitions.  S denotes SMILES and F denotes SELFIES;
Supplementary Section~7 reports the complete three-block table.}
\label{tab:v48-causal-main}
\end{table}

Positive correspondence improves over the LM on both SMILES--ZINC
($\Delta=-.0321$, 95\% CI $[-.0571,-.0051]$) and SELFIES--ZINC
($-.6049$, $[-.6890,-.5218]$).  Shuffling removes the gain, while the tested
symmetric-KL variant does not improve either block.  Thus chemically correct
hidden-state correspondence, not additional computation, is the shared
effective ingredient.  The margin adds a clear SMILES benefit
($-.0505$, $[-.0774,-.0226]$ versus the LM), but not a universal one.
Supplementary Section~7 reports the complete contrasts and training costs.

\subsection{Chemical Information Is Preserved}

Covariance effective rank and descriptor-probe $R^2$ increase on all four
SMILES domains, while fingerprint macro-AUROC remains within .004 of canonical
training.  Scaffold-conditioned similarities retain between-molecule
separation, arguing against representational collapse.  Complete values appear
in Supplementary Section~8.

\subsection{Molecular Property Prediction}

Table~\ref{tab:downstream-replay-main} shows that EOS-level SIGMA improves the
mean on all six property tasks, with four
improving in every seed.  ESOL and Lipophilicity intervals exclude zero; the
others are favorable but inconclusive.  Variation across two strings for the
same molecule decreases on every task (0.2--44.9\%); complete intervals and
per-seed results appear in Supplementary Section~5.
\section{Discussion and Limitations}

Exact-suffix certification controls the comparison, dense supervision shapes
the prediction path, and the finite margin preserves non-equivalent states.
Position accuracy, ordering, next-token agreement, likelihood, and effective
rank connect the design to its local mechanism.  SMILES--ChEMBL and SAFE show
that local consistency cannot substitute for global closure or fragment
assembly.

SIGMA requires certifiable same-suffix triplets, so coverage varies.  Absolute
SMILES and SELFIES FCD values are not directly comparable.  Pair construction
adds training-time cost but leaves inference unchanged; larger backbones,
broader corpora, and laboratory utility remain future work.

The certificate uses Standard-InChIKey equivalence, not Fixed-H or exact
charged-graph identity, so mobile-H tautomers and some charge drawings can
share an identity.  A graph-exact contract would require rebuilding triplets
and re-evaluating the models.
\section{Conclusion}

SIGMA turns molecular serialization equivalence into dense supervision through
certified same-suffix triplets.  One shared setting improves FCD across four
SELFIES domains and SMILES--QM9/ZINC; mechanism controls identify correct
hidden-state correspondence as the effective ingredient; EOS-level transfer
improves six property benchmarks.  The framework makes representation
equivalence an actionable training signal.

\clearpage
\bibliography{references}

\end{document}